\documentclass[pdflatex,sn-basic]{sn-jnl}% Basic Springer Nature Reference Style/Chemistry Reference Style
\usepackage{graphicx}%
\usepackage{multirow}%
\usepackage{amsmath,amssymb,amsfonts}%
\usepackage{amsthm}%
\usepackage{mathrsfs}%
\usepackage[title]{appendix}%
\usepackage{xcolor}%
\usepackage{textcomp}%
\usepackage{manyfoot}%
\usepackage{booktabs}%
\usepackage{algorithm}%
\usepackage{algorithmicx}%
\usepackage{algpseudocode}%
\usepackage{textcomp, graphicx, graphics, soul, url, epsfig, balance, microtype}%
\usepackage{listings}%
\usepackage{inconsolata}
\usepackage{textcomp, graphicx, graphics, soul, url, epsfig, balance, microtype}
\usepackage{amsmath,amssymb,amsfonts}
\usepackage{xcolor}
\usepackage{hyperref}
\usepackage{subcaption}
\usepackage{multirow, booktabs}
\usepackage{xcolor}
\theoremstyle{thmstyleone}%
\theoremstyle{thmstyletwo}%

\theoremstyle{thmstylethree}%

\begin{document}

\title[Article Title]{IRWOZ 2.0: A Large Language Model-driven Dialogue Dataset for Industrial Robot Conversations}


\author*[1]{\fnm{Chen} \sur{Li}}\email{cl@mp.aau.dk}

\author[1]{\fnm{Dimitrios} \sur{Chrysostomou}}\email{dimi@mp.aau.dk}

\affil*[1]{\orgdiv{Department of Materials and Production}, \orgname{Aalborg University}, \orgaddress{\street{Fibigerstraede 16}, \city{Aalborg}, \postcode{9220}, \country{Denmark}}}


\abstract{IRWOZ has improved industrial human-robot interaction (HRI) dialogue systems through domain-specific annotations. However, its initial version contains substantial noise in dialogue states and utterances, limiting state-tracking accuracy. We introduce IRWOZ 2.0, which addresses these limitations through large language model (LLM) enhanced generation (Mistral/Claude-3.5) and quality refinements. Our improved dataset expands to 390 dialogues across 4 industrial domains (Assembly, Delivery, Position, Relocation), featuring manual corrections and automated typo removal. Benchmark experiments on dialogue state tracking demonstrate significant improvements, with GPT-2's BLEU-4 score increasing from 0.1651 to 0.5604 compared to original IRWOZ. To support industrial HRI research, we publicly released IRWOZ 2.0 dataset at https://ieee-dataport.org/documents/irwoz-20-large-language-model-driven-dialogue-dataset-industrial-robot-conversations}


\maketitle

\section{Introduction}\label{sec:intro}
\begin{figure}[t]
    \centering
    \includegraphics[width=0.8\textwidth]{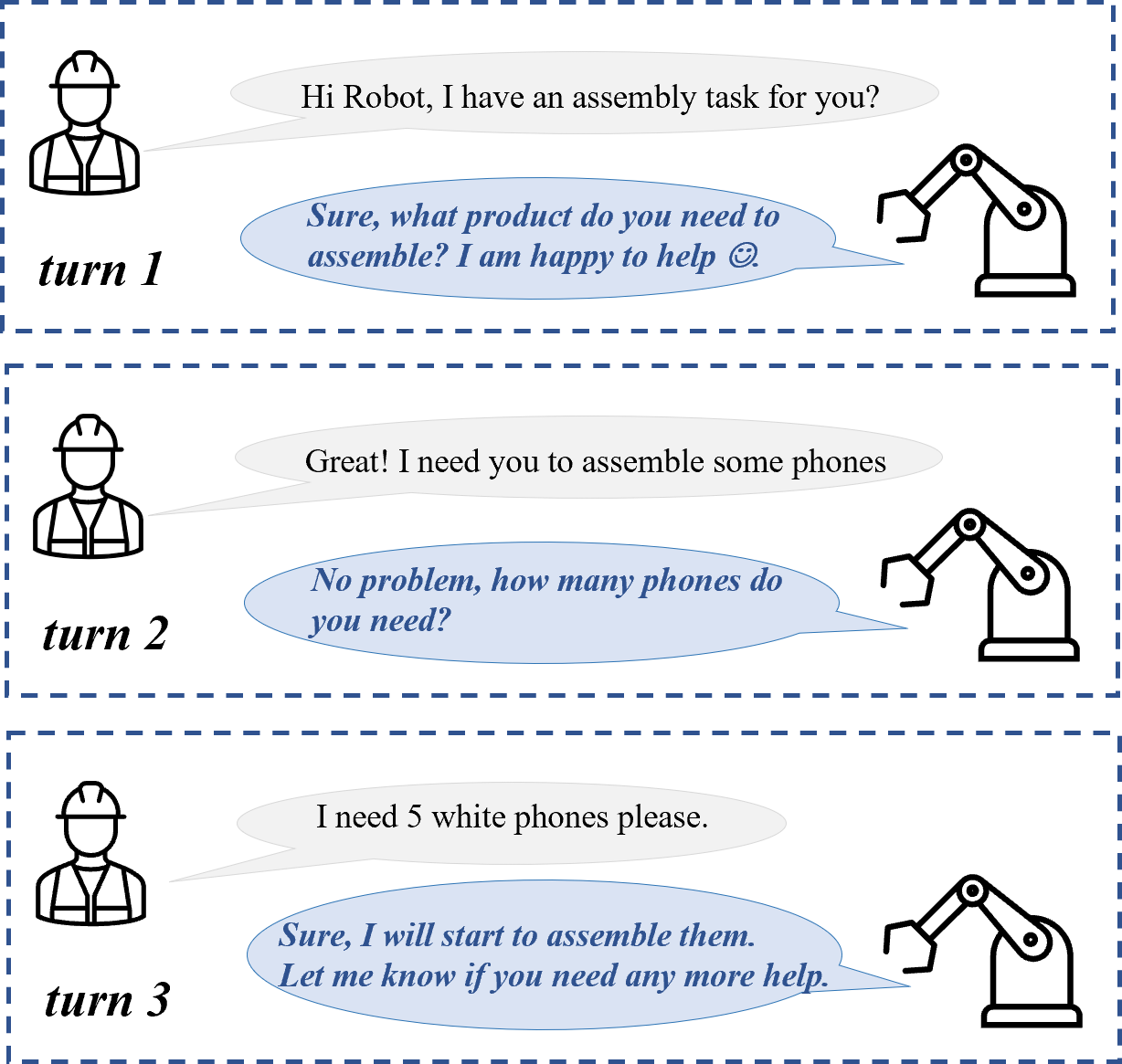}
    \caption{Example dialogue for phone assembly tasks in industrial settings, demonstrating operator-robot interaction sequences.}
    \label{fig:scenario}
\end{figure}

The rapid development of artificial intelligence (AI) has led to its increased adoption across industries, with human-robot interaction (HRI) playing a significant role in enhancing productivity and efficiency~\cite{campagna2024promoting}. Natural language-enabled AI agents show particular potential for revolutionizing HRI by facilitating seamless communication between humans and robots~\cite{li2023speech, zheng2024review}. However, developing such agents requires high-quality dialogue datasets that accurately capture industrial complexities~\cite{kernan2023harnessing, ni2023recent}. Figure~\ref{fig:scenario} illustrates a typical three-turn dialogue between a human operator and industrial robot during phone assembly tasks.\par
Traditional dialogue dataset creation relies on manual collection and annotation through Wizard-of-Oz (WoZ) experiments. This approach presents substantial challenges in industrial environments, where data collection proves both time-consuming and costly. Capturing diverse industrial scenarios remains particularly challenging, while multi-annotator involvement often introduces inconsistencies that limit scalability for comprehensive domain coverage~\cite{irwoz}.\par
Our previous IRWOZ dataset~\cite{irwoz} represents significant progress in addressing these challenges. However, manual collection and annotation processes still introduce substantial noise in both annotations and utterances, negatively impacting dialogue state tracking and response generation performance~\cite{yi2024survey}. Recent analyses indicate error rates exceeding 20\% in manually annotated industrial dialogues~\cite{yi2024survey}.\par
To address these limitations, we present IRWOZ 2.0 - a systematic framework combining large language models (LLMs) with three key innovations for industrial dialogue dataset enhancement: (1) automated LLM-powered error detection/correction, (2) structured prompt engineering for domain-specific generation, and (3) hybrid validation integrating automated checks with expert review. Our implementation leverages Mistral~\cite{jiang2023mistral} and Claude-3.5~\cite{anthropic_claude} to expand the original dataset to 390 dialogues across four critical industrial domains (Assembly, Delivery, Position, Relocation), achieving 3.1$\times$ faster generation than manual methods while preserving technical accuracy. Benchmark results demonstrate this framework's effectiveness through GPT-2's improved BLEU-4 scores for dialogue state tracking (0.1651 → 0.5604), alongside maintained domain specificity in complex industrial scenarios. The contribution of the paper is as follows:
\begin{itemize}
    \item \textbf{Enhanced Data Quality}: Implementation of hybrid quality control measures (automated and manual) reducing annotation errors while improving dialogue state tracking reliability
    
    \item \textbf{Expanded Domain Coverage}: Comprehensive coverage across four industrial domains (Assembly, Delivery, Position, Relocation) with 390 dialogues capturing diverse interaction scenarios
    
    \item \textbf{LLM Integration}: Systematic framework combining Mistral~\cite{jiang2023mistral} and Claude-3.5~\cite{anthropic_claude} for efficient, context-aware dialogue generation
    
    \item \textbf{Rigorous Evaluation}: Comprehensive benchmarking and human evaluation protocols assessing performance improvements over IRWOZ
\end{itemize}

\section{Related Work}
\label{sec:related}
In this section, we review three key areas relevant to our work: task-oriented dialogue datasets, the use of LLMs for dataset correction, and LLMs for dataset generation. These areas form the foundation for our approach in creating IRWOZ 2.0, a novel dataset for industrial robot conversations. 

\subsection{Task-Oriented Dialogue Datasets}
Task-oriented dialogue datasets play a crucial role in advancing research in dialogue systems~\cite{reimann2024survey}. Traditional pipeline models for task-oriented dialogue systems often require explicit modeling of dialogue states and hand-crafted action spaces to interact with domain-specific knowledge bases~\cite{wen2018sequence}. Task-oriented dialogue systems aim to assist users in completing specific tasks such as booking flights, making restaurant reservations, or providing information. Traditional datasets for such systems are often created through manual annotation or by collecting real-world interactions. For instance, the ATIS (Airline Travel Information System) dataset~\cite{hemphill1990atis} has been a benchmark for spoken language understanding systems, containing flight information queries. 

Another significant dataset is the MultiWOZ dataset~\cite{budzianowski2018multiwoz}, which is a large-scale multi-domain dialogue dataset containing interactions spanning over multiple domains and services. These datasets have been instrumental in advancing the field by providing structured and annotated data for training and evaluating dialogue systems. More recently, the Schema-Guided Dialogue (SGD) dataset~\cite{rastogi2020towards} has been introduced, which includes over 20,000 dialogues spanning 16 domains. SGD aims to provide a more comprehensive and flexible framework for task-oriented dialogue systems by incorporating schema information that describes the entities and actions within each domain. The Frames dataset~\cite{el2019frames} is another example, which focuses on the challenge of handling implicit information in task-oriented dialogues, providing a testbed for systems to infer and act upon such information. 

However, even though these datasets have significantly advanced dialogue systems research, they primarily focus on general domains such as travel and hospitality. The industrial robotics domain, with its unique vocabulary and task-specific interactions, remains underrepresented. This gap underscores the need for specialized datasets like IRWOZ 2.0 that can capture the nuances of HRI in industrial settings.

\subsection{LLMs for Dataset Correction}
LLMs have become pivotal in the realm of dataset corrections, effectively addressing a spectrum of issues such as typographical errors within dialogue datasets. With their deep understanding of language structure and context, LLMs are adept at detecting and rectifying typographical errors, ensuring the dataset's text is clean and coherent. CoachLM represents a significant approach, automatically enhancing the quality of instruction pairs in datasets~\cite{coachlm2024}. The model's effectiveness is demonstrated by its real-world application at Huawei, where it increased dataset cleaning efficiency by 20\%. Studies also explore LLMs' self-correction abilities, indicating that external feedback and large-scale fine-tuning bolster their reliability~\cite{survey2024}. Additionally, GrammarGPT has shown promise in correcting grammatical errors in native Chinese text using a hybrid dataset~\cite{grammargpt2023}. In software engineering, LLMs predict implicit dataflows in dynamically typed code for dataflow graph generation, outperforming traditional static analysis methods~\cite{dataflow2023}. However, applying LLMs to correct specialized datasets such as those for industrial robotics presents unique challenges, including the need to preserve domain-specific terminology and task-relevant information while improving overall quality.

\subsection{LLMs for Dataset Generation}
The use of LLMs has revolutionized the way dialogue datasets are generated. LLMs, such as GPT~\cite{radford2018improving} and BERT~\cite{devlin2018bert}, have been fine-tuned on dialogue data to generate realistic and coherent conversational data. These models can produce large volumes of dialogue data that can be used for training and testing dialogue systems. For example,~\cite{adiwardana2020towards} introduced a model that generates responses indistinguishable from human-written ones, showcasing the potential of LLMs in creating high-quality dialogue datasets by enabling the generation of diverse and nuanced dialogue scenarios.\par
In addition to generating new datasets, LLMs have been used to enhance existing datasets by augmenting them with more diverse and contextually relevant responses. For instance, the work by~\cite{zhang2021conversational} explores the use of LLM for text-to-text generation task, to improve conversational machine reading datasets, making them more robust and adaptable to real-world applications. Furthermore, the DialoGPT model~\cite{zhang2020dialogpt} represents a significant step forward in generating conversational responses, demonstrating the ability of LLMs to capture the nuances of human dialogue.\par 
It is evident, that the traditional datasets offer the advantage of domain-specific knowledge and structured annotations, however, they often require significant manual effort and are limited in size and diversity. On the other hand, LLM-generated datasets provide scalability and a broader range of conversational contexts but may lack the depth of domain-specific knowledge. Our work with IRWOZ 2.0 represents a hybrid approach, leveraging LLMs to generate scalable, diverse, and industry-relevant dialogue datasets while maintaining focus on specific industrial domains. This method is particularly important for advancing dialogue systems capable of handling the complex demands of real-world industrial applications, an area where existing LLM applications have been limited.\par
\section{Dataset Corrections}
\label{sec:dataset}
To create IRWOZ 2.0, we first addressed the limitations and errors present in the original IRWOZ dataset. The original IRWOZ dataset was collected using a Wizard-of-OZ setup among 18 participants, including students, academics, and shop floor workers with backgrounds ranging from robotics to computer science. The participants were randomly paired as two to construct dialogues. One participant played the role of the Wizard and the other plays the shop floor worker. The shop floor worker is asked to choose a task and initiate a dialogue in one of domains, \textit{Delivery}, \textit{Position}, \textit{Assembly} and \textit{Relocation}. The wizard needed to respond to the shop floor worker according to the required task. The web application was constructed with separate interfaces, one for the user and one for the Wizard. The wizard was given access to the back-end database and a list of the robot-controlling APIs to verify the resource requested by user.\par 
Regardless the great effort to develop such web application for dialogue dataset collection and annotation, along with implementing human validation processes, there are many mistakes in IRWOZ such as incomplete markups, and mis-annotations. These errors arise from misinterpretations, inconsistencies among annotators, and the complexities of contextual nuances in dialogues. Such inaccuracies can significantly affect the quality of the dataset and the performance of dialogue systems.\par
\begin{figure*}[t]
    \centerline{\includegraphics[width=1.0\textwidth]{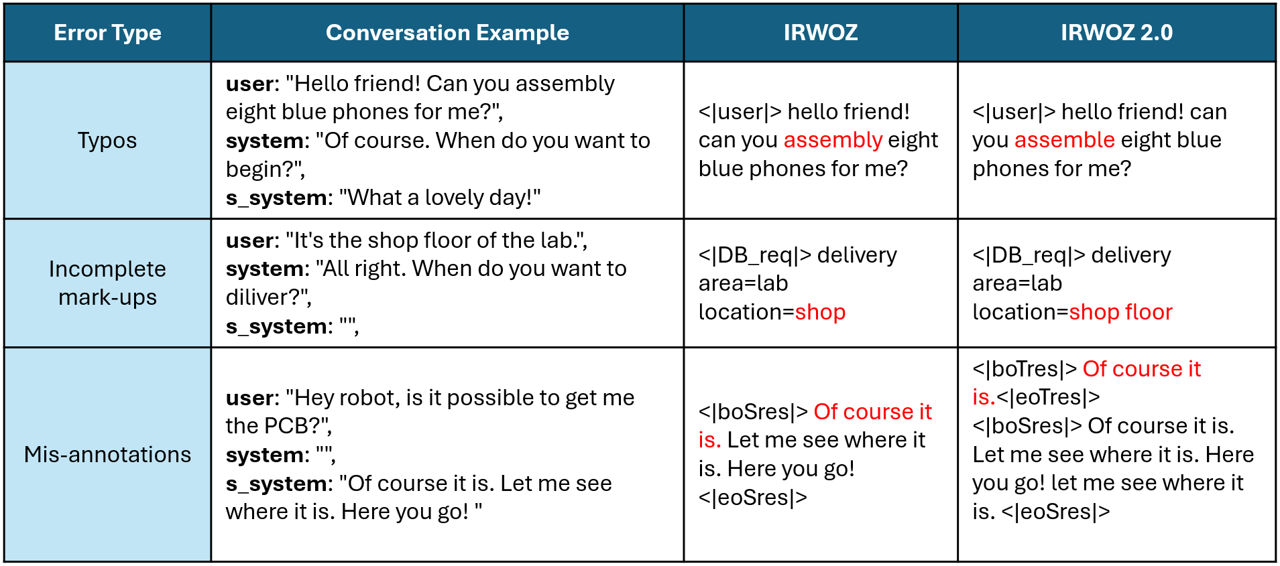}}
    \caption{Examples of annotation errors between IRWOZ and IRWOZ 2.0 ($<|DB\_req|>$, $<|user|>$, $<|boSres|>$, $<|eoSres|$, $<|boTres|>$ and $<|eoTres|>$ are special tokens defined to annotate the dialogue dataset~\cite{irwoz}.}
    \label{fig:error}
\end{figure*}
\subsection{Error Types}
The most common errors types in the original dialogue dataset:
\begin{itemize}
    \item Typos. The text, including user utterances, responses, and slots, contains typographical errors that can affect its accuracy and effectiveness. For example, in Figure~\ref{fig:error}, the phrase "can you \textit{assembly} 8 blue phones for me?" The word "assembly" should be replaced by "assemble." 
    \item Incomplete mark-ups. The slot value was not fully extracted from the user's input. This frequently happens with compound nouns. For instance, in Figure~\ref{fig:error}, we observe this scenario in which  the term "shop floor" denotes the physical location of the laboratory, yet only "shop" is extracted from the user's utterance.   
    \item Mis-annotations. The value is assigned to an incorrect slot type. For example, in Figure~\ref{fig:error}, the robot's response to the shop floor work, "Let me see where it is," is intended to be task related. However, it is incorrectly marked as small talk.
\end{itemize}
\subsection{Dialogue Utterance Corrections}
Dialog dataset correction was done through Manual corrections and Automated corrections with the help of the LLM, Claude-3.5. 

\textbf{Manual Corrections}. Manual corrections were employed to address mis-annotations and incomplete markups identified within the dataset. This process involved two steps:
\begin{itemize}
    \item \textit{Slot Values Verification}: verification of each slot value was made to determine if it was correctly marked against the corresponding user utterance.
    \item \textit{Response Verification}: check if the given utterance was annotated against the dialogue context and against the task requirements.
\end{itemize}
These manual corrections were essential for maintaining the integrity of the dataset, especially in cases where nuanced understanding and contextual judgement were required. This process was particularly crucial for preserving the accuracy of domain-specific terminology and task-related information unique to industrial robotics scenarios. The Table~\ref{tab:corr} shows the statistical analysis results of the mis-annotations and incomplete markups, i.e., 2.2\% of the turns have incomplete markups and 3.2\% of the turns have incorrect annotations. 

\textbf{Automated Corrections}.To enhance the efficiency and accuracy of the correction process, automated corrections were implemented using Claude-3.5. This LLM was specifically fine-tuned to recognize and preserve industrial terminology while correcting typographical errors. The following steps were involved in the automated correction process:
\begin{itemize}
    \item \textit{Prompt Definition}: A specific prompt (see Appendix ~\ref{sec:prom_correct}) was defined to guide Claude-3.5 in detecting and correcting typos in the dialogue dataset.
    \item \textit{Execution}: The IRWOZ dataset were uploaded as a separate document, where Claude-3.5 processed the dataset according to the defined prompt, systematically identifying and correcting typographical errors.
\end{itemize}
The Table~\ref{tab:corr} shows the statistical analysis results of typos in the IRWOZ, i.e., 4.3\% of the turns contain typos or misspellings. The correction process significantly improved the overall quality of the dataset. By reducing typos, incomplete markups, and mis-annotations, we had substantial improvements in the performance of dialogue systems trained on this data. This is particularly important in industrial settings where miscommunication between humans and robots can lead to inefficiencies or safety issues.
\begin{table}[t]
    \centering
    \renewcommand{\arraystretch}{1.2}
        \begin{tabular}{@{}lc@{}}
        \toprule
        Error Type & \% of errors \\
        \midrule
        Typos and Misspellings & 4.3\% \\
        Incomplete Markups & 2.2\% \\
        Mis-annotations & 3.2\% \\
        \bottomrule
    \end{tabular}
    \caption{Statistical analysis of manual and automated corrections of the IRWOZ dataset.}
    \label{tab:corr}
\end{table}
\section{Dataset Generation}
\label{sec:dataset_gen}
Even though the IRWOZ was collected through WoZ method, which allows for rich, realistic interactions by simulating system responses with human operators, the collection and annotation of such dialogue datasets for HRI in industrial domains is a labor-intensive and time-consuming task. Therefore, we leverage the LLMs' capabilities to generate realistic and contextually relevant dialogues that reflect the language-enabled HRIs in four industrial domains. 
\begin{figure*}[t]
  \centering
    \includegraphics[width=1.0\linewidth]{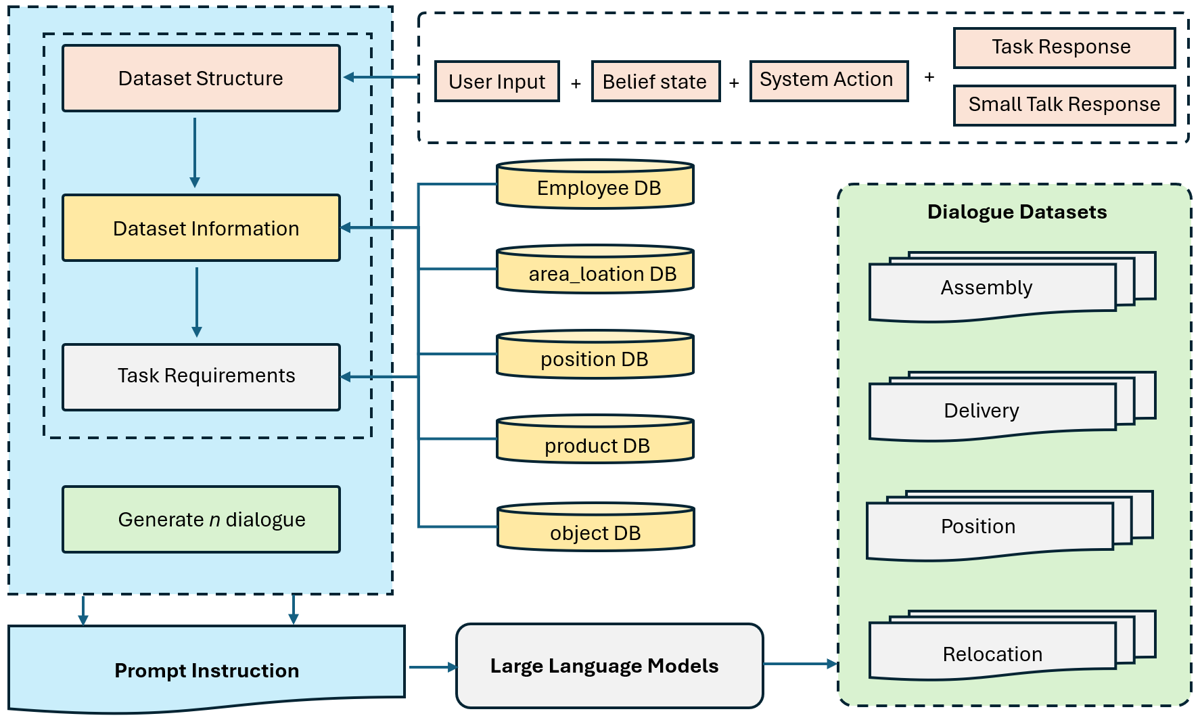}
  \caption{The overall architecture of the dialogue generation framework for building the IRWoZ 2.0 dataset.}
  \label{fig:framework}
\end{figure*}
\subsection{Overall Framework}
Figure~\ref{fig:framework} illustrates the overall architecture of the dialogue generation framework for generating the IRWOZ 2.0 dataset. The framework's foundation lies in the detailed design of task-specific prompts for each industrial domain. The framework is composed of three components:
\begin{itemize}
    \item \textbf{Prompt Instructions}. It includes 1) dataset structure definition including user input, dialogue belief state, system action, and corresponding system responses, including both task-oriented and small talk responses; 2) database information,  facilitating accurate generation and verification of belief states and system actions, from key database tables (e.g., area, product) incorporated into the prompt, and 3) task requirements specifying behavioral constraints for the language models, such as the protocol for informing users about non-existent products in the database.
    \item \textbf{LLMs}: Claude-3.5 and Mistral, are leveraged to generated the datasets based on the given prompt instructions.
    \item \textbf{Dialogue Datasets}: generated dialogues for each industrial domain (i.e., Delivery, Position, Assembly, and Relocation). 
\end{itemize}
\begin{figure*}[t]
    \centerline{\includegraphics[width=1.0\textwidth]{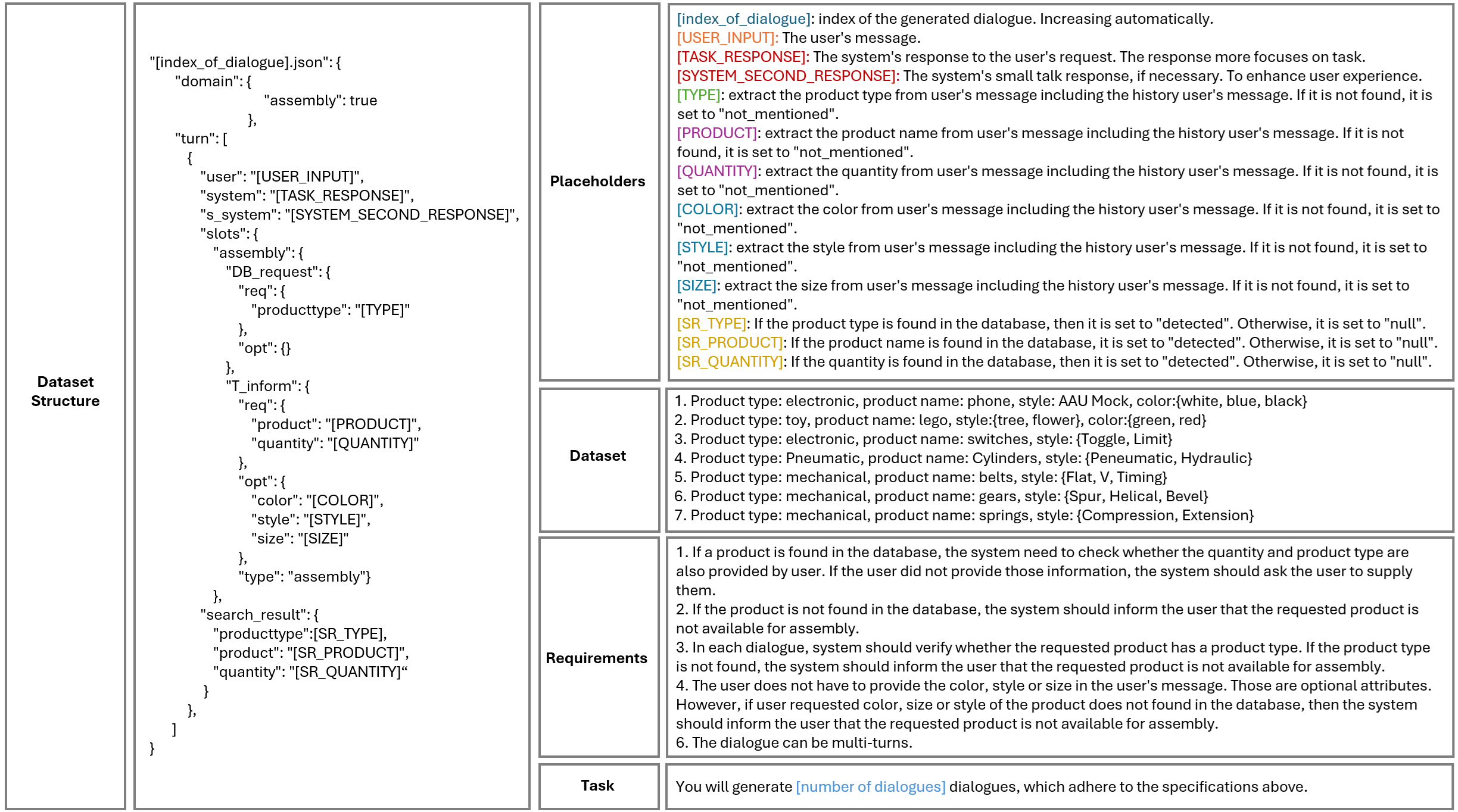}}
    \caption{An example of designed prompt for the Assembly task for Claude-3.5 and Mistral}
    \label{fig:prompt}
\end{figure*}
\subsection{Prompt Design}
While LLMs are highly effective in handling conversations, the quality and style of their responses are largely dependent on the prompts they receive from users. Well-set up and articulate prompts are imperative for LLMs to exhibit the desired behavior.\par 
The primary objective is to ensure the generated dialogue dataset is contextually relevant and technically accurate. This requires a clear definition of the industrial task which the prompt is trying to address — a detailed background of how the task is usually carried out as well as the specific roles of humans and robots in the corresponding scenario. As an example, a prompt designed to generate a dialogue for a relocation task would need to specify: the items to be relocated, where to relocate them to, and any barriers that could be run into. The prompt should also encourage the generation of dialogues that incorporate common industrial terminologies and commands, ensuring that the robot's responses are precise and aligned with the industry domain.\par 
Furthermore, the prompt is designed to simulate realistic HRI by incorporating natural language elements such as disambiguation, confirmation, and error-handling. This requires creating scenarios in which the human operator may ask the robot to report the status of the task, verify a correct action, or fix any problem during the task. For example, the generated response for the relocation task like “Has the PCB been correctly put on the conveyor belt?” or "What to do if the PCB cannot be found in the desired location." By embedding such interactions in the conversations, the generated dialogues are able to address the operational aspects of the tasks, enhance the robot's ability to communicate effectively, and respond to unexpected situations. Moreover, due to the fact that LLMs are trained on vast and diverse text data across various contexts, they can effectively mimic human speech styles and generate human-like responses. This capability aligns with one of the key objectives of the IRWOZ dataset: to provide both task-related and small talk responses, thereby enhancing the user experience during HRI.\par 
Our optimal prompt for the Assembly task is shown in Figure~\ref{fig:prompt}. The prompt begins with a required dataset structure of the desired dialogue. The special tokens from IRWOZ~\cite{irwoz} are leveraged to annotate the dialogue context, dialogue belief, system actions and system response. Each of the placeholder, which is annotated with colored '[]',is given the corresponding explanation. It is then followed by an example from previous IRWOZ dataset to provide extra information to instruct the dataset generation process. We use the enumerated list to point out all the requirements that constrain the LLM's behavior. When we provide only the high level goal instruction to Claude-3.5 and Mistral, we use the term  "Task". Similar prompts are designed for the other three tasks, Relocation, Deliver and Position. Appendix ~\ref{sec:gen_sample} shows two generated dialogue samples for the Assembly task using Claude-3.5 and Mistral.\par 
\subsection{Data Statistics}
A total of 390 dialogues were generated and annotated across four industrial domains: Assembly, Delivery, Position, and Relocation. The distribution of dialogue turns for each domain in Appendix ~\ref{sec:turn_dis}. Each of the above domain has around 54\%-55\% 1 turn dialogues, 33\%-35\% 2 turns dialogues and only around 10\% -11\% are 3 turns dialogues. Which makes sense since keeping short and concrete dialogues for work related HRI is major happens in shop floor. 
\begin{table}[!t]
    \centering
    \renewcommand{\arraystretch}{1.1}
    \begin{tabular}{@{}llccc@{}}
        \toprule
        \textbf{Task} & \textbf{Role} & \textbf{Min} & \textbf{Max} & \textbf{Avg} \\
        \midrule
        Assembly & User & 3 & 10 & 6.98 \\
        & Task Resp. & 9 & 48 & 23.29 \\
        & ST Resp. & 4 & 22 & 12.90 \\
        \addlinespace
        Delivery & User & 3 & 14 & 8.91 \\
        & Task Resp. & 8 & 62 & 26.71 \\
        & ST Resp. & 0 & 22 & 10.30 \\
        \addlinespace
        Position & User & 4 & 21 & 8.22 \\
        & Task Resp. & 6 & 47 & 25.41 \\
        & ST Resp. & 2 & 26 & 11.96 \\
        \addlinespace
        Relocation & User & 6 & 24 & 10.77 \\
        & Task Resp. & 5 & 56 & 21.35 \\
        & ST Resp. & 3 & 21 & 11.67 \\
        \bottomrule
    \end{tabular}
    \small
    \caption{Sentence length (words) distribution in IRWOZ 2.0. Task Resp.: Task-related responses, ST Resp.: Small talk responses}
    \label{tab:sen_len}
\end{table}

% \begin{table}[!ht]
%     \centering
%     \renewcommand{\arraystretch}{1.2}
%     \begin{tabular}{@{}lllccc@{}}
%         \toprule
%         \textbf{Task} & \textbf{Role} & \textbf{Min} & \textbf{Max} & \textbf{Average} \\
%         \midrule
%         \multirow{3}{*}{Assembly} 
%             & User utterance & 3 & 10 & 6.98 \\
%             & $T\_Response$ & 9 & 48 & 23.29 \\
%             & $ST\_Response$ & 4 & 22 & 12.90 \\
%         \midrule
%         \multirow{3}{*}{Delivery}
%             & User utterance & 3 & 14 & 8.91 \\
%             & $T\_Response$ & 8 & 62 & 26.71 \\
%             & $ST\_Response$ & 0 & 22 & 10.30 \\
%         \midrule
%         \multirow{3}{*}{Position}
%             & User utterance & 4 & 21 & 8.22 \\
%             & $T\_Response$ & 6 & 47 & 25.41 \\
%             & $ST\_Response$ & 2 & 26 & 11.96 \\
%         \midrule
%         \multirow{3}{*}{Relocation}
%             & User utterance & 6 & 24 & 10.77 \\
%             & $T\_Response$ & 5 & 56 & 21.35 \\
%             & $ST\_Response$ & 3 & 21 & 11.67 \\
%         \bottomrule
%     \end{tabular}
%     \small
%      \caption{Sentence length (words) distribution of IRWOZ 2.0 ($T\_Response$: task related system response, $ST\_Response$: small talk related response)}
%      \label{tab:sen_len}
% \end{table}
%
Table~\ref{tab:sen_len} illustrates the distribution of sentence length for user utterances, task-related responses, and small talk responses. The average length of user utterances varies across domains, ranging from 6.98 words for Assembly to 10.77 words for Relocation. Task-related responses are generally longer, with averages ranging from 21.35 words for Relocation to 26.71 words for Delivery. Small talk responses are consistently shorter across all domains, with averages between 10.30 and 12.90 words. As expected, each domain contains system responses without small talk involved. The varied response lengths contribute to improved generalization ability for training models. The maximum sentence length for task-related responses reaches up to 62 words in the Delivery domain, while user utterances and small talk responses have lower maximum lengths, enhancing the dataset's diversity and realism. \par
\begin{figure}[!ht]
    \centering
    \subfloat[User utterance length distribution]{
        \includegraphics[width=0.6\textwidth]{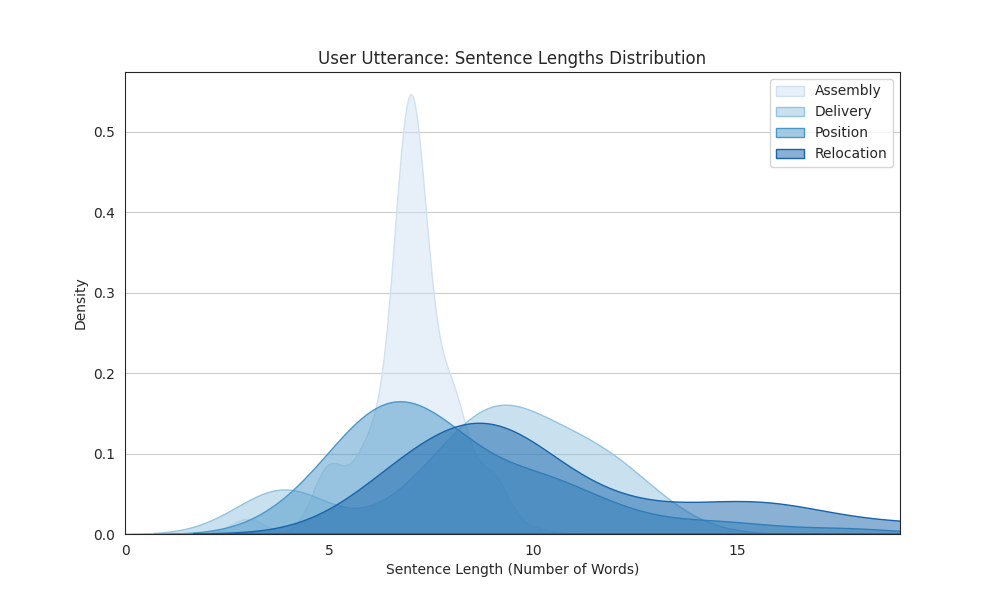}
        \label{fig:subfigA}
    }
    
    \vspace{0.1cm}  % 子图之间的垂直间距
    
    \subfloat[Task response length distribution]{
        \includegraphics[width=0.6\textwidth]{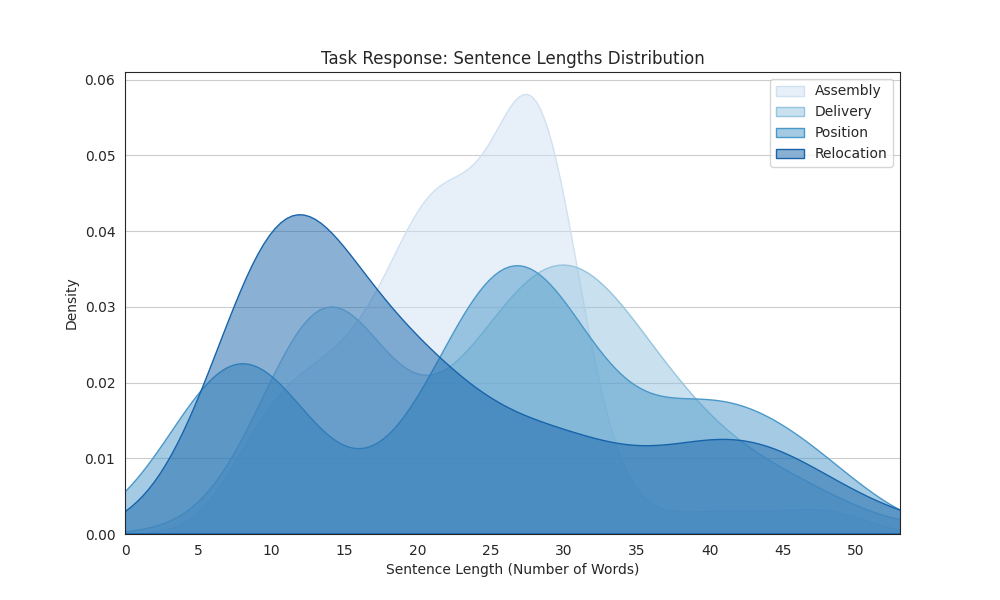}
        \label{fig:subfigB}
    }
    
    \vspace{0.1cm}
    
    \subfloat[Small talk responses length distribution]{
        \includegraphics[width=0.6\textwidth]{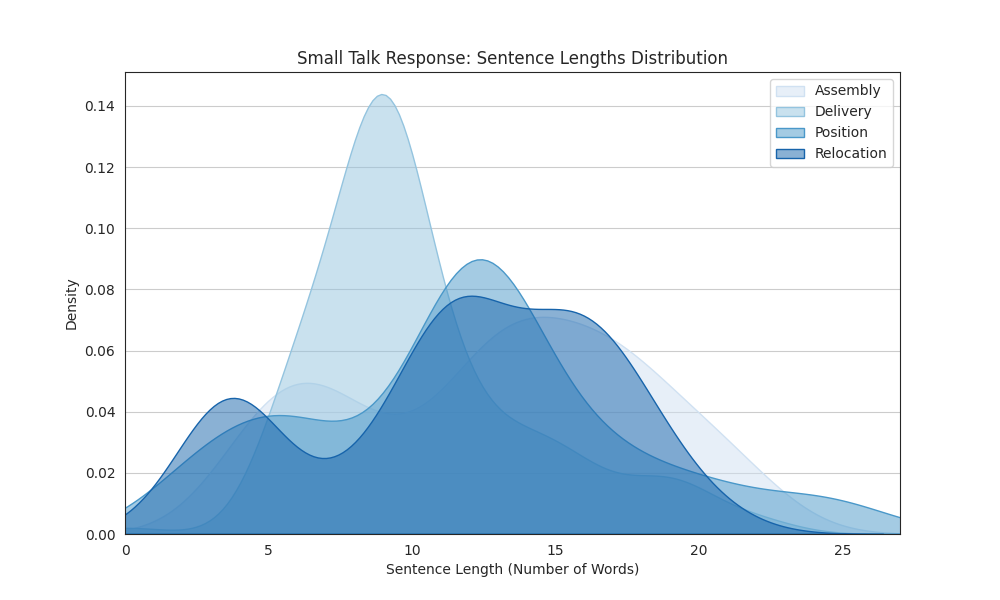}
        \label{fig:subfigC}
    }
    \caption{Distribution of sentence lengths across four industrial domains (Assembly, Delivery, Position, and Relocation) in IRWOZ 2.0.}
    \label{fig:sen_len}
\end{figure}
Figure~\ref{fig:sen_len} illustrates varied sentence length distributions across four domains. User utterances show task-specific ranges, with Relocation the widest (6-24 words) and Assembly the narrowest (3-10 words). Task responses consistently have the broadest distributions, peaking at 62 tokens for Delivery, indicating detailed explanations in industrial settings. Small talk responses are briefer (0-26 words) and more consistent across domains. Notably, Delivery exhibits the most variation overall, suggesting higher complexity. These patterns reflect diverse communication needs in industrial dialogues, with task responses requiring the most linguistic flexibility, while small talk remains concise across all domains.\par 
\begin{table*}[!ht]
    \centering
    \renewcommand{\arraystretch}{1.2}
    \setlength{\tabcolsep}{4.5pt}   % 默认约6pt，压缩到4.5pt
    \footnotesize                    % 比 small 稍小一点，但比 tiny 大很多
    \begin{tabular}{@{}llccccccc@{}}
        \toprule
        \textbf{Model} & \textbf{Dataset} & \textbf{Perplexity} & \textbf{BLEU-1} & \textbf{BLEU-2} & \textbf{BLEU-3} & \textbf{BLEU-4} & \textbf{SA} & \textbf{JGA} \\
        \midrule
        \multirow{2}{*}{GPT-2}
            & IRWOZ     & \textbf{1.05} & 0.3816 & 0.2680 & 0.2088 & 0.1651 & \textbf{0.949} & 0.801 \\
            & IRWOZ 2.0 & 1.10 & \textbf{0.7370} & \textbf{0.6581} & \textbf{0.6059} & \textbf{0.5604} & 0.944 & \textbf{0.824} \\
        \midrule
        \multirow{2}{*}{GPT-2 Medium}
            & IRWOZ     & \textbf{1.05} & 0.6214 & 0.5620 & 0.5352 & 0.5114 & \textbf{0.979} & \textbf{0.925} \\
            & IRWOZ 2.0 & 1.11 & \textbf{0.7392} & \textbf{0.6633} & \textbf{0.6114} & \textbf{0.5656} & 0.956 & 0.855 \\
        \midrule
        \multirow{2}{*}{GPT-2 Large}
            & IRWOZ     & \textbf{1.05} & 0.6493 & 0.5949 & 0.5681 & 0.5446 & \textbf{0.983} & \textbf{0.938} \\
            & IRWOZ 2.0 & 1.10 & \textbf{0.7399} & \textbf{0.6645} & \textbf{0.6134} & \textbf{0.5681} & 0.960 & 0.871 \\
        \bottomrule
    \end{tabular}
    \\ \textit{Note:} BLEU (bilingual evaluation understudy), SA (Slot Accuracy), and JGA (Joint goal accuracy).
    \caption{Performance comparison of three different model architectures on IRWoZ and IRWOZ 2.0 dataset}
    \label{tab:100dataset}
\end{table*}

\section{Benchmark Evaluation}
\label{sec:benchmark}
\subsection{Benchmark Models}
To ensure a fair comparison between the proposed IRWOZ 2.0 and its predecessor, experiments are conducted using dialogues generated by LLMs from IRWOZ 2.0, allowing for a comparison with the original IRWOZ dataset. We also employed the same evaluation methodology using three GPT-2 language models: gpt2, gpt2-medium, and gpt2-large ~\cite{irwoz}. This comparison aims to quantify any performance differences and assess whether LLM-generated datasets can match or potentially surpass the quality of manually curated ones in various dialogue system tasks.

\subsection{Results and Discussion}
As we can see in Table~\ref{tab:100dataset}, one of the most notable improvements with IRWOZ 2.0 is the substantial increase in BLEU scores across all gpt2 model variants. IRWOZ 2.0 consistently outperforms the original IRWOZ dataset in BLEU metrics for all model sizes. For instance, the gpt2 model exhibits a BLEU-1 score increase from 0.3816 (IRWOZ) to 0.7370 (IRWOZ 2.0), and BLEU-4 from 0.1651 to 0.5604. Similar improvements are observed in gpt2-medium and gpt2-large models, with BLEU-4 scores reaching 0.5656 and 0.5681, respectively, compared to their IRWOZ counterparts. These enhancements indicate that IRWOZ 2.0 facilitates the generation of more fluent, relevant, and contextually appropriate dialogues, which are crucial for realistic HRIs. 

IRWOZ 2.0 demonstrates notable improvements in joint goal accuracy (JGA), particularly with the base gpt2 model.The gpt2 model’s JGA increases from 0.801 (IRWOZ) to 0.824 (IRWOZ 2.0), showcasing a better ability to fulfill dialogue objectives. Although gpt2-medium and gpt2-large show higher JGA with the original IRWOZ, the base gpt2 model benefits significantly from IRWOZ 2.0, suggesting that even smaller models can achieve enhanced performance with the improved dataset. 

IRWOZ generally exhibits slightly better perplexity scores across all model sizes, even if the differences are marginal (IRWOZ: 1.05 vs. IRWOZ 2.0: 1.10 for gpt2, IRWOZ 2.0: 1.11 for gpt2-medium, and IRWOZ 2.0: 1.10 for gpt2-large). Given that BLEU scores and JGA are more indicative of dialogue quality and goal achievement, the slight increase in perplexity is outweighed by the substantial gains in language generation and goal accuracy. Though the original IRWOZ dataset generally shows higher slot accuracy (SA) scores, IRWOZ 2.0 maintains competitive performance. For the gpt2 model, SA remains high (0.944 for IRWOZ 2.0 vs. 0.949 for IRWOZ). Although gpt2-medium and gpt2-large experience a slight dip in SA with IRWOZ 2.0 (0.956 and 0.960 respectively) compared to IRWOZ (0.979 and 0.983), the overall SA scores are still commendable, ensuring reliable intent recognition and response generation. 

Although some metrics like Perplexity show minor trade-offs, the substantial improvements in BLEU scores and JGA underscore the dataset's enhanced language generation capabilities and effectiveness in achieving dialogue objectives. These positive outcomes affirm that IRWOZ 2.0 is a valuable asset for training more sophisticated, responsive, and reliable HRI in industrial settings. 
\begin{figure}[!t]
  \centering
    \includegraphics[width=.8\linewidth]{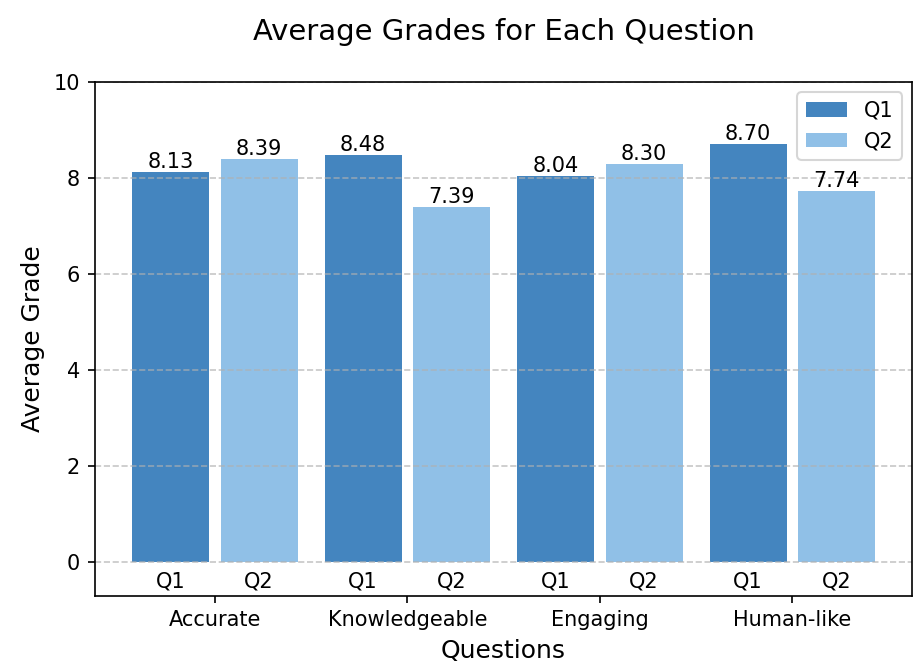}
  \caption{The score distribution of different raters}
  \label{fig:grade}
\end{figure}
\section{Human Evaluation}
\label{sec:humeva}
We also perform a human evaluation on the quality of the LLM generated dataset. We randomly sampled 40 dialogues (see Appendix ~\ref{sec:hum_evl}) from each domain and recruited 23 participants from different nationalities and background, including shop floor workers with industrial robot operation experience ($\geq 3$ years), robotics engineers, computational linguistics expert, robotics/AI researchers and students and industrial automation researchers. Specifically, we followed the similar strategy from ~\cite{ye-etal-2022-multiwoz} and ~\cite{adding}, the raters were asked to assign a score from 1 to 10 for each question (see Appendix ~\ref{sec:hum_evl}) based on the following criteria: 1) Accurate: the generated dialogue demonstrates accurate understanding of context and extracting the correct slots from user's request; 2) Knowledgeable: the dialogues contains meaningful and relevant industrial scenarios; 3) Engaging: the dialogue maintains natural interaction flow and encourages human-robot collaboration; 4) Human-like: the interaction feels natural while maintaining appropriate robot-specific characteristics.\par
%
% The evaluation results demonstrate strong performance in accuracy metrics (Accurate Q2: 8.60, Q1: 8.06) particularly in contextual understanding of user intents, reflecting the LLM's capability to maintain coherent task-oriented interactions. This aligns with findings from \cite{ye-etal-2022-multiwoz} regarding the importance of precise slot-value annotation in dialogue systems. The knowledgeable dimension (Q1: 8.31, Q2: 7.06) reveals domain-specific expertise in industrial scenarios while indicating opportunities for improved integration of technical details during task execution phases. Engagement scores (Q1: 8.31, Q2: 7.81) suggest effective turn-taking mechanics but reveal challenges in maintaining dynamic interaction flows during complex multi-step operations. Human-like characteristics show the highest single-question score (Q1: 8.50) in politeness maintenance, while lower performance in language variation (Q2: 7.62) indicates room for improvement in generating diverse natural expressions without compromising technical precision.\par
The evaluation results (see Figure ~\ref{fig:grade}) demonstrate strong performance in accuracy metrics (Accurate Q1: 8.13, Q2: 8.39) particularly in contextual understanding of user intents, reflecting the LLM's capability to maintain coherent task-oriented interactions. This aligns with findings from \cite{ye-etal-2022-multiwoz} regarding the importance of precise slot-value annotation in dialogue systems. The knowledgeable dimension (Q1: 8.48, Q2: 7.39) reveals domain-specific expertise in industrial scenarios while indicating opportunities for improved integration of technical details during task execution phases. The engagement scores (Q1: 8.04, Q2: 8.30) suggest effective turn-taking mechanics but reveal challenges in maintaining dynamic interaction flows during complex multi-step operations. The human-like characteristics show the highest single-question score (Q1: 8.70) in politeness maintenance, while performance in language variation (Q2: 7.74) indicates room for improvement in generating diverse natural expressions.\par 
%
%
% This score suggests the evaluation criteria effectively capture essential aspects of human-robot interaction quality, similar to the annotation consistency improvements reported in \cite{ye-etal-2022-multiwoz}. Participant feedback highlighted particular strengths in safety protocol discussions and maintenance troubleshooting scenarios, while noting occasional over-formalization in conversational turns during equipment calibration sequences.\par
%
\section{Conclusions \& Future Work}
\label{sec:conclusion}
In this paper, we present IRWOZ 2.0, an advanced dialogue dataset for industrial HRI. It combines LLM-based generation with manual and automated correction processes, resulted in a more comprehensive and accurate dataset covering four key industrial domains: Assembly, Delivery, Position, and Relocation. Benchmarking results show that IRWOZ 2.0 outperforms its predecessor, with notable improvements in BLEU scores, Joint Goal Accuracy, and other key metrics across various model architectures. The public release of IRWOZ 2.0 will provide the research community with a valuable resource for developing more sophisticated and responsive dialogue systems for industrial HRI. Future work may focus on expanding to additional industrial domains, integrating multimodal data, and exploring transfer learning techniques to further improve model performance across various industrial scenarios.\par

\bmhead{Author Contributions} Author Chen Li prepared the initial manuscript, conducted data generation, and benchmark evaluation. Dimitrios Chrysostomou contributed to data correction and data analysis. All authors have reviewed, contributed to, and approved of the manuscript.

\bmhead{Data availability}
Our IRWOZ dataset is available at https://github.com/lcroy/ToD4IR/tree/main/dataset, the IRWOZ 2.0 dataset is available for non-commercial purpose at: IEEE dataport: https://ieee-dataport.org/documents/irwoz-20-large-language-model-driven-dialogue-dataset-industrial-robot-conversations.

\section*{Declarations}
\bmhead{Conflict of interest} All authors declare that they have no known competing financial interests or personal relationships that could have appeared to influence the work reported in this paper.

\bibliography{reference}

\begin{appendices}

\section{Prompt for correction}
\label{sec:prom_correct}
\begin{figure}[h]
    \centerline{\includegraphics[width=0.4\textwidth]{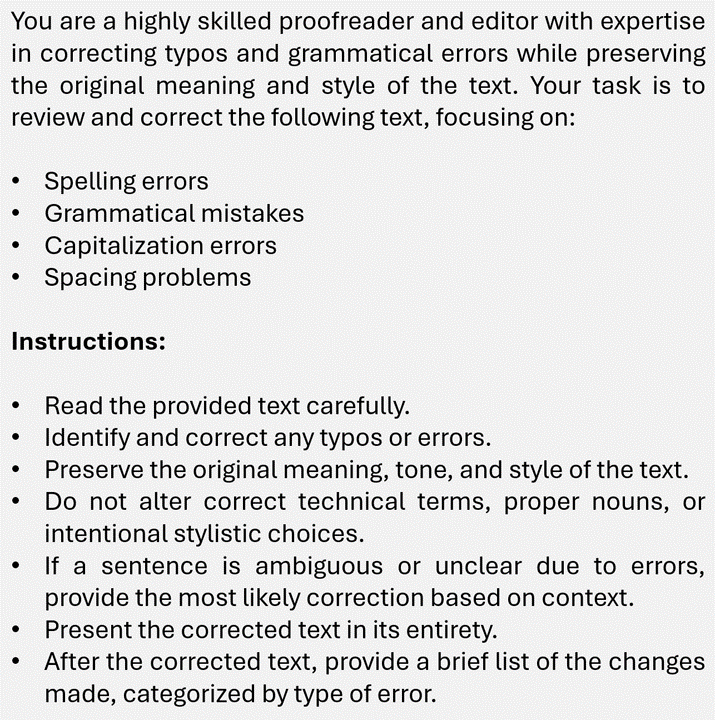}}
    \caption{The defined prompt for remove the typos from the IRWOZ dataset}
    \label{fig:correction}
\end{figure}
\section{Dialogue turn distribution of each domain}
\label{sec:turn_dis}
\begin{figure*}[h]
    \centering
    \subfloat[Domain: Assembly]{
        \includegraphics[width=0.4\textwidth]{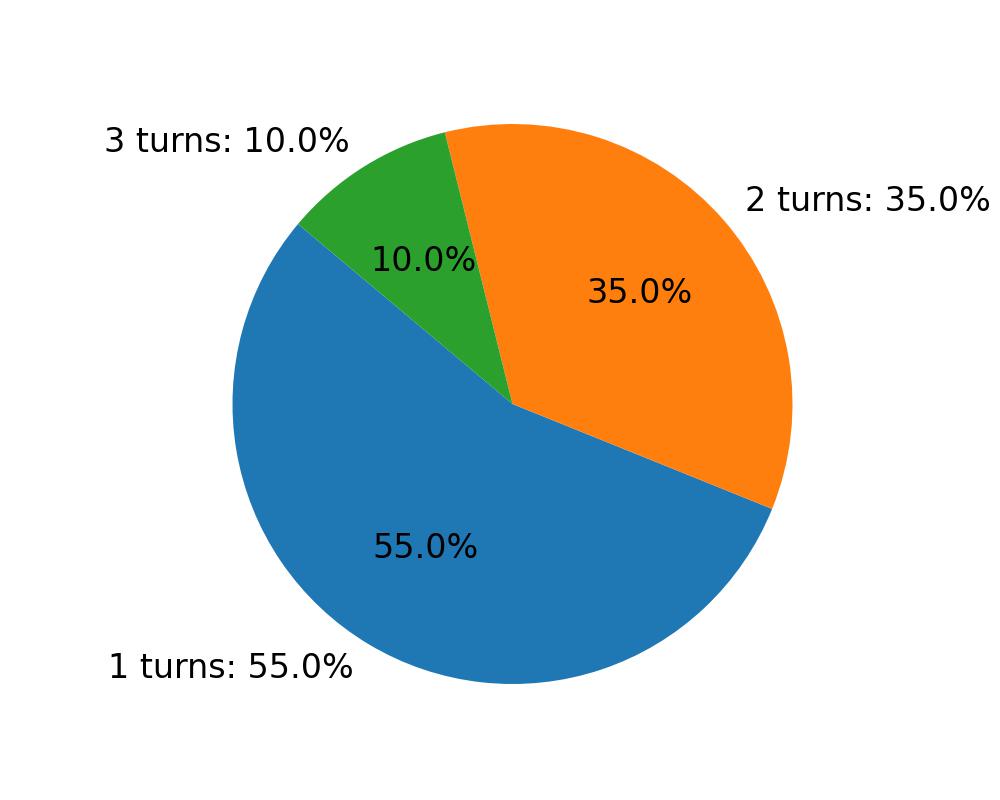} % Replace with your image file
        \label{fig:subfig1}
    }
    \hfill
    \subfloat[Domain: Delivery]{
        \includegraphics[width=0.4\textwidth]{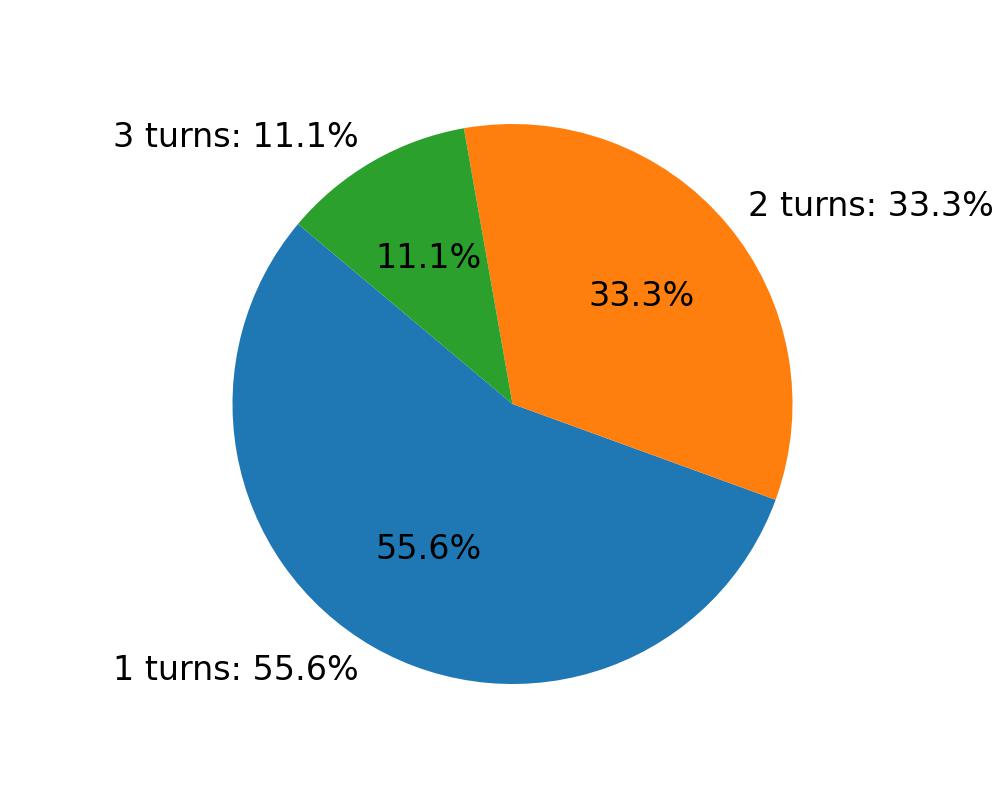} % Replace with your image file
        \label{fig:subfig2}
    }
    
    \vspace{0.5cm} % Space between rows

    \subfloat[Domain: Position]{
        \includegraphics[width=0.4\textwidth]{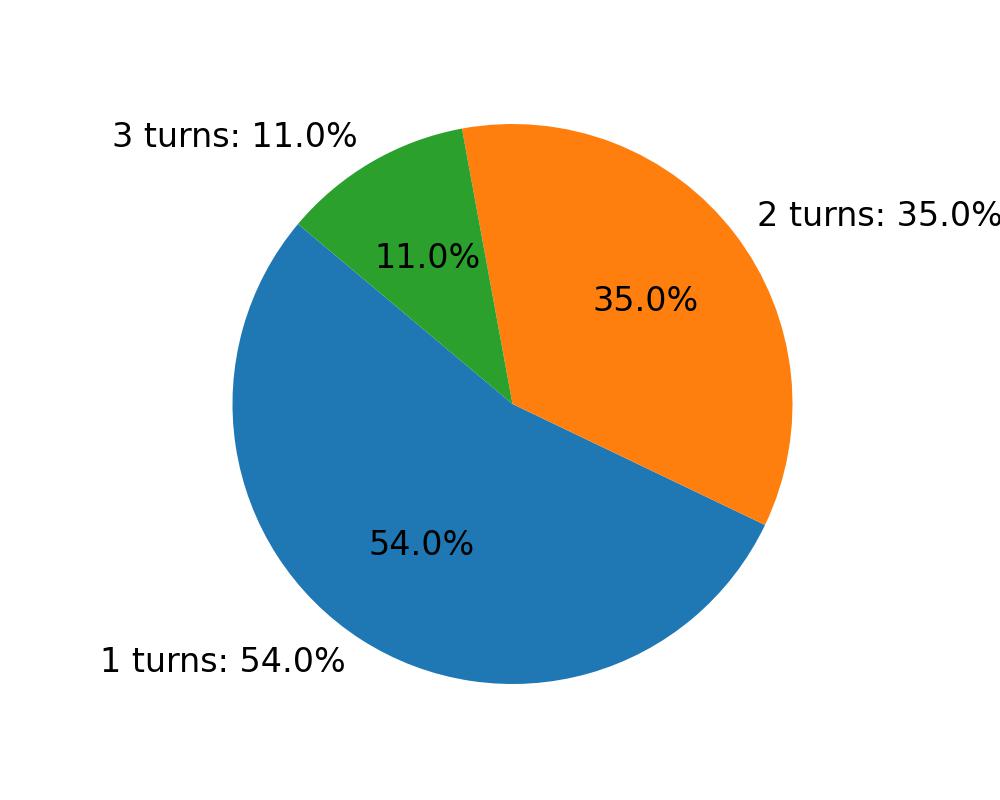} % Replace with your image file
        \label{fig:subfig3}
    }
    \hfill
    \subfloat[Domain: Relocation]{
        \includegraphics[width=0.4\textwidth]{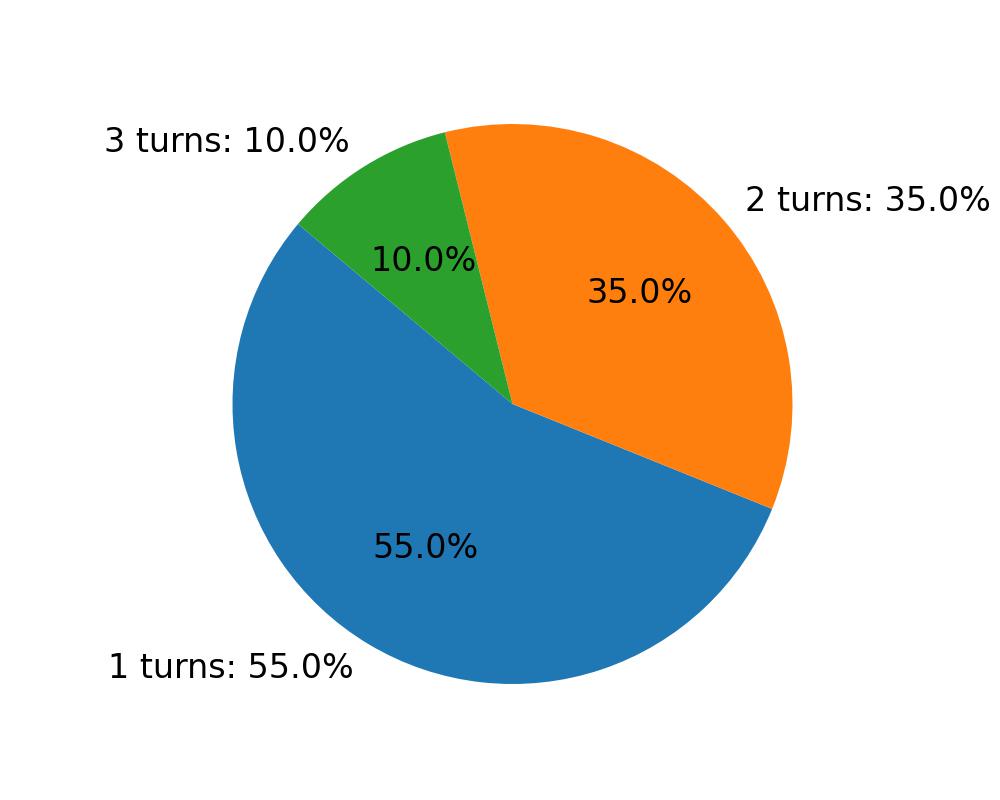} % Replace with your image file
        \label{fig:subfig4}
    }
    \caption{Dialogue turn distribution of each domain}
    \label{fig:turn_domain}
\end{figure*}

Figure~\ref{fig:turn_domain} shows the generated dialogues turn distribution of four domains. 
\begin{figure*}[!t]
    \centerline{\includegraphics[width=1.0\textwidth]{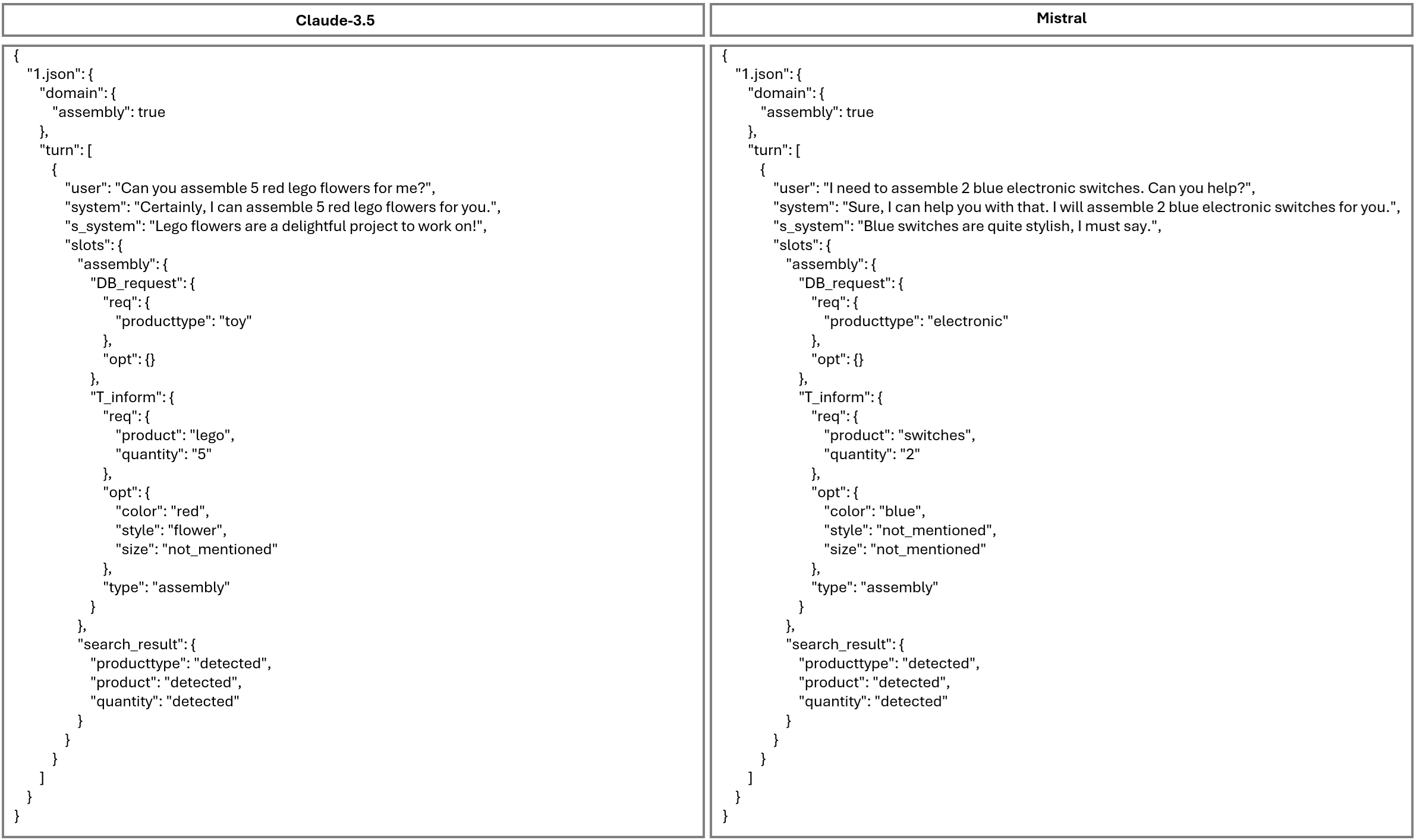}}
    \caption{Two generated dialogue samples for the Assembly task using Claude-3.5 and Mistral}
    \label{fig:sample}
\end{figure*}

\begin{table*}[!t]
\centering
\footnotesize                      % 从 small 降到 footnotesize
\renewcommand{\arraystretch}{1.2}
\setlength{\tabcolsep}{4pt}        % 压缩列间距
\begin{tabular}{p{1.8cm}p{5.5cm}cccccccc}  % 略微收窄前两列
\toprule
\textbf{Category} & \textbf{Question} & \textbf{G3} & \textbf{G4} & \textbf{G5} & \textbf{G6} & \textbf{G7} & \textbf{G8} & \textbf{G9} & \textbf{G10} \\
\midrule
\multirow{2}{*}{Engaging} 
& Q1: The dialogue flow maintains a good pace without long delays or abrupt endings & -- & -- & 3 & 1 & 4 & 5 & 4 & 6 \\
& Q2: The dialogue includes proper turn-taking and interaction dynamics & -- & -- & 2 & 1 & 8 & 4 & 3 & 6 \\
\midrule
\multirow{2}{*}{Knowledgeable}
& Q1: The dialogue covers the relevant four industrial & -- & -- & 2 & 1 & 2 & 4 & 7 & 7 \\
& Q2: The dialogue maintains coherence throughout and includes appropriate task-specific details & 2 & 1 & -- & 3 & 2 & 9 & 4 & 2 \\
\midrule
\multirow{2}{*}{Humanlike}
& Q1: The generated responses show appropriate levels of politeness & 1 & -- & -- & 2 & 1 & 3 & 6 & 10 \\
& Q2: The dialogue includes natural language variations and expressions & -- & 1 & 1 & 2 & 4 & 9 & 3 & 3 \\
\midrule
\multirow{2}{*}{Accurate}
& Q1: The dialogue demonstrates accurate understanding of the user's intent & -- & 1 & 1 & 1 & 5 & 3 & 7 & 5 \\
& Q2: The responses show accurate understanding of the context of the conversation & -- & -- & 1 & -- & 3 & 10 & 3 & 6 \\
\bottomrule
\end{tabular}
\caption{Results of human evaluation on dialogue quality. Each cell shows the number of human evaluators who gave that grade (G3-G10 represent grades from 3 to 10, where '--' indicates zero responses). Grades 1 and 2 are not shown in the table as no evaluators assigned these grades. The evaluation was conducted across four categories with two questions each, measuring different aspects of dialogue quality.}
\label{tab:evaluation_matrix}
\end{table*}

\section{Generated Dialogue Samples}
\label{sec:gen_sample}
Figure~\ref{fig:sample} shows two generated dialogue samples for the Assembly task.
\section{Human Evaluation - Questions and Results}
\label{sec:hum_evl}
\begin{figure*}[b]
    \centering
    \centerline{\includegraphics[width=1.0\textwidth]{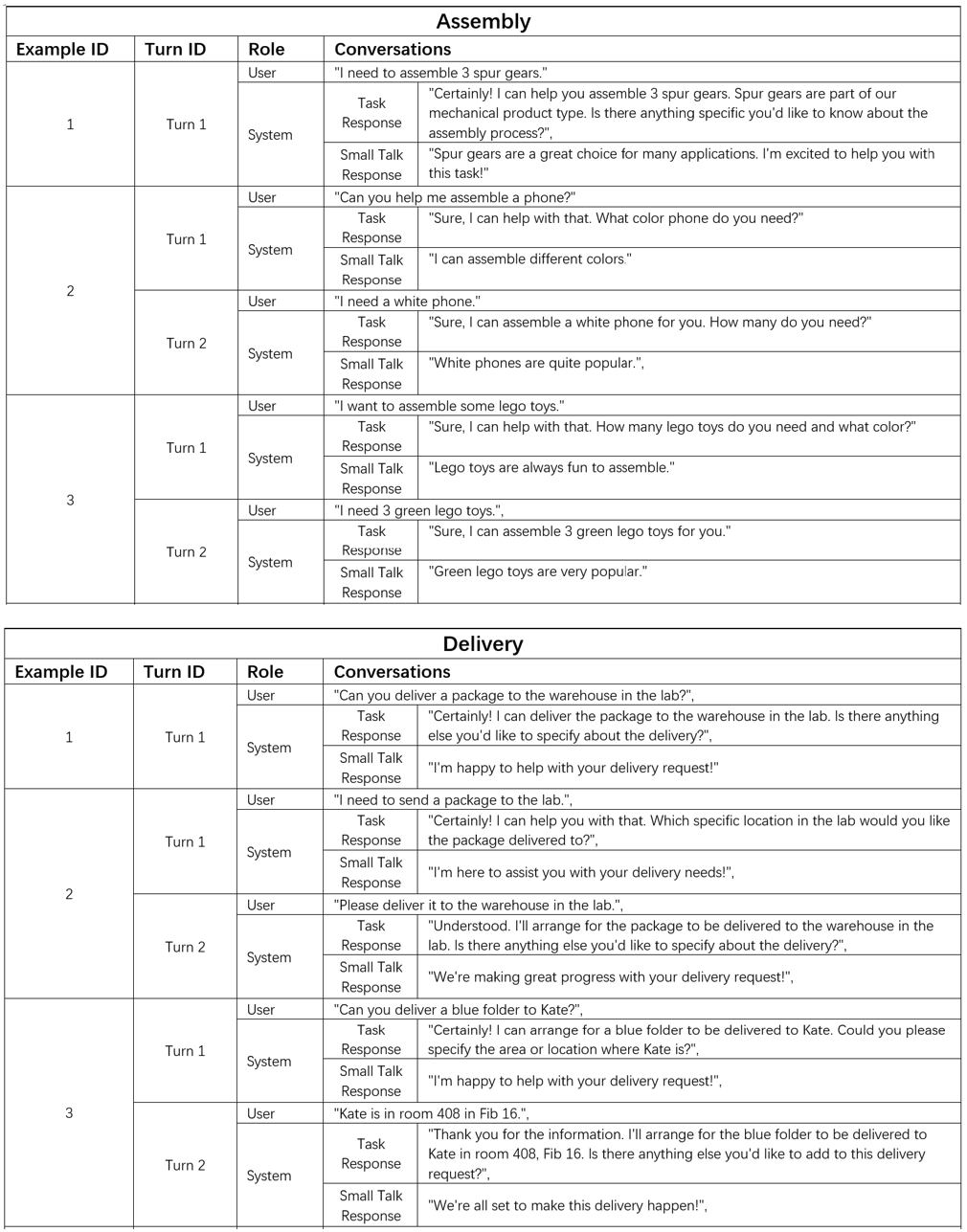}}
    \caption{Examples of randomly selected LLM-generated dialogues from Assembly and Delivery domains}
    \label{fig:samp_1}
\end{figure*}

\begin{figure*}[b]
    \centering
    \centerline{\includegraphics[width=1.0\textwidth]{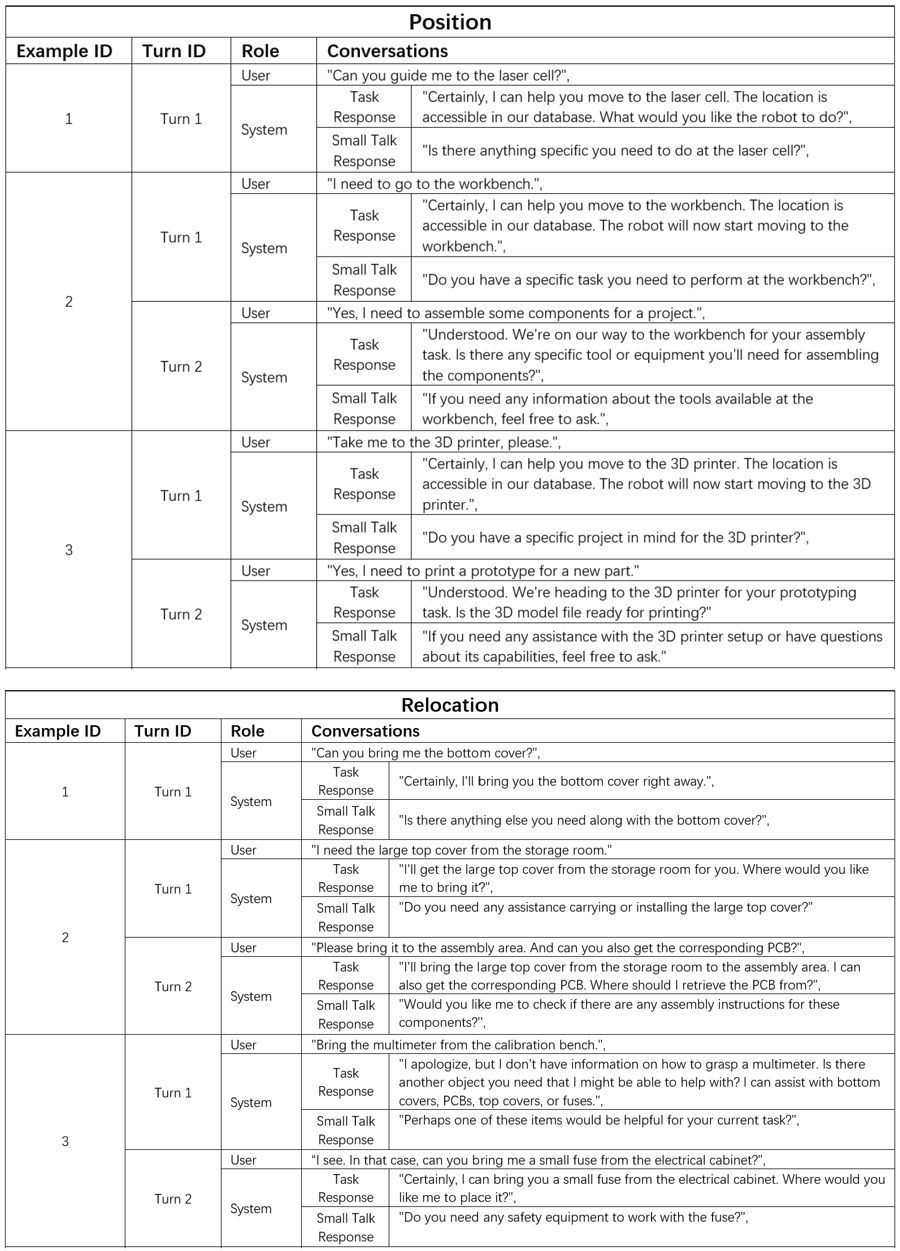}}
    \caption{Examples of randomly selected LLM-generated dialogues from Position and Relocation domainsains}
    \label{fig:samp_2}
\end{figure*}

Table ~\ref{tab:evaluation_matrix} shows the results of human evaluation on dialogue quality. Figure ~\ref{fig:samp_1} and ~\ref{fig:samp_2} show examples of randomaly selected LLM-generated dialogues for human evaluation.

\end{appendices}


\end{document}